\documentclass[runningheads]{llncs}
\usepackage{graphicx}
\usepackage{amsmath}

\usepackage{multirow}
\usepackage{makecell}

\usepackage{hyperref}

\begin{document}
\title{Instantaneous Physiological Estimation using Video Transformers}

%
%

\author{%
Ambareesh Revanur\inst{1}\thanks{Corresponding Author: \email{arevanur@andrew.cmu.edu} \\ URL: \url{https://github.com/revanurambareesh/instantaneous_transformer} 
\\ Preprint. Accepted at \textit{Springer - Studies in Computational Intelligence 2022}
} \quad %
Ananyananda Dasari\inst{2}\quad%
Conrad S. Tucker\inst{2}\quad%
\quad\quad L\'aszl\'o A. Jeni\inst{1}
}

\authorrunning{Revanur et al.}
%
\institute{Robotics Institute, Carnegie Mellon University \\
\and
Dept. of Mechanical Engineering, Carnegie Mellon University}
\maketitle              
\begin{abstract}
Video-based physiological signal estimation has been limited primarily to predicting episodic scores in windowed intervals. 
While these intermittent values are useful, they provide an incomplete picture of patients' physiological status and may lead to late detection of critical conditions. 
We propose a video Transformer for estimating instantaneous heart rate and respiration rate from face videos. Physiological signals are typically confounded by alignment errors in space and time. To overcome this, we formulated the loss in the frequency domain. 
We evaluated the method on the large scale Vision-for-Vitals (V4V) benchmark. It outperformed both shallow and deep learning based methods for instantaneous respiration rate estimation. In the case of heart-rate estimation, it achieved an instantaneous-MAE of 13.0 beats-per-minute. 
\end{abstract}

\section{Introduction}

\noindent Contact-based devices (e.g. pulse-oximeter) are prevalent among healthcare professionals for assessing and monitoring the vital signs of patients in hospital settings. These vital sign monitoring devices require physical contact and can cause discomfort to patients.  As remote diagnosis is becoming increasingly common, partly due to the recent COVID-19 pandemic, there is a pressing demand for non-contact physiological estimation methods.

Over the years, conventional photoplethysmography (PPG), the contact-based optical estimation of microvascular blood volume changes, has evolved into contactless imaging PPG (iPPG). These methods utilize digital cameras and computer vision techniques for estimating heart-generated pulse waves and their respiratory modulation by means of peripheral blood perfusion measurements. Past research has demonstrated that these bio-signals can be extracted with high fidelity in a strictly controlled environment; i.e. with a prediction error $<3$ beats-per-min for Heart Rate extraction \cite{eccv2018deepphys}. Intuitively, the camera captures subtle periodic color variations that result from the blood volume changes in the underlying skin tissues. A careful analysis of the subtle changes in the video reveals the physiological state.

\begin{figure}[t]
    \centering    \includegraphics[width=0.65\columnwidth]{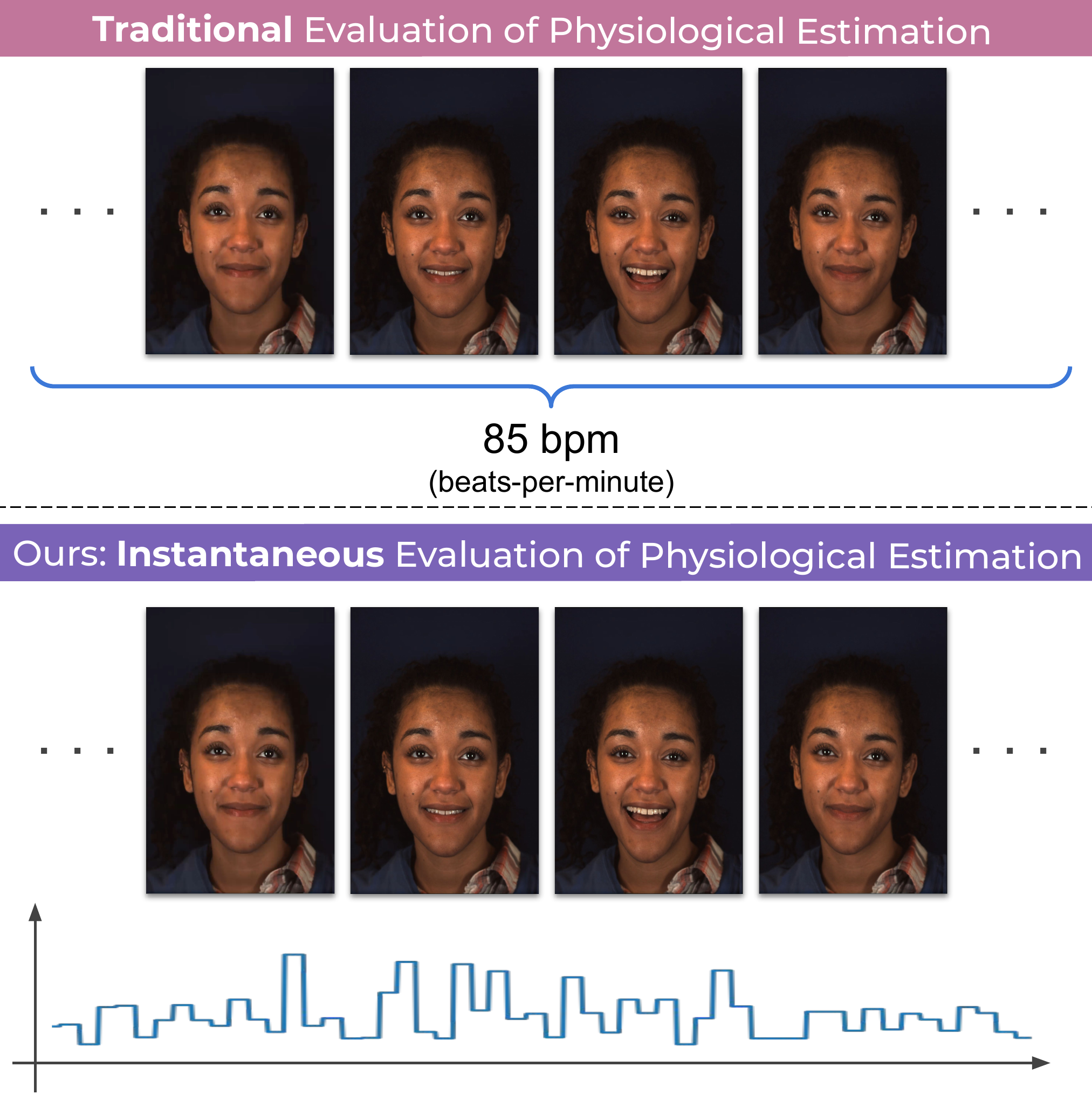}
    \caption{Traditional evaluation v/s  instantaneous evaluation (Ours) for heart rate and respiration rate estimation}
    \label{fig:concept}

\end{figure}

Previous video-based physiological extraction techniques have been limited to predicting episodic heart rate values over large, \textit{non-overlapping} windows (e.g. 30 seconds), which influenced the choice of performance metrics to evaluate these methods. The window-based evaluation protocol does not provide complete insight into Heart Rate Variability (HRV), which plays an important role in understanding the physical and mental conditions of an individual \cite{npjafib2020}. This evaluation gap has been considered in a recent physiological challenge organized at the ICCV conference called ``Vision-for-Vitals" (V4V). Instead of using non-overlapping windows to measure the error, the challenge organizers \cite{revanur2021first} proposed using an instantaneous (a.k.a. \textit{continuous})  evaluation protocol that measures the performance of a method at a per-frame level. Hence, in our work, we aim to benchmark different methods including our proposed method using this continuous evaluation protocol (Fig. \ref{fig:concept}).

Recently, deep learning based solutions have been proposed for the task of human physiology estimation \cite{eccv2018deepphys,eccv_20_video}. One of the most popular methods in this direction is DeepPhys \cite{eccv2018deepphys}. This method utilizes the optical principles of PPG to predict blood volume pulse (and respiratory wave for respiration rate estimation) from a facial video. Even though DeepPhys has made significant progress, it still limited to episodic evaluation and is unable to fully exploit the temporal and periodic nature of the blood volume pulse. To remedy this, we draw motivation from recent work on video analysis \cite{neimark2021video} and utilize a Transformer based architecture for frame-level prediction. There is a body of literature that shows that the Transformer is effective in modelling temporal sequences \cite{vaswani2017attention,carion2020end}.

The paper advances two main novelties:

\begin{itemize}
    \item In our work we use a Transformer based architecture for instantaneous prediction and evaluation of human physiological signals from the facial video. We formulate the loss function in the frequency domain to minimize confounds introduced by the temporal misalignment of the video and the PPG signals.
    
    \item We evaluate the method on the challenging Vision-for-Vitals (V4V) dataset. We show that our approach reliably estimates Heart Rate (HR) and  Respiration Rate (RR), outperforming shallow and deep-learning methods trained on the same data  (V4V training set). 
\end{itemize}

\section{Related Work}

\subsection{Video based physiology extraction}
Based on the principles of remote-photoplethysmography (rPPG), several methods have been advanced \cite{Wu12Eulerian,green2008remote,wang2016algorithmic,ica2010non,chrom2013robust,npjBiases2021,bkf,tarassenko2014non3in1} for the extraction of physiological signals from facial videos. In \cite{green2008remote}, the authors highlighted that the green channel of the RGB video can be used to compute the heart rate since the green channel has the strongest signature of photoplethysmography. Wang et. al \cite{wang2016algorithmic} proposed a mathematical model for the reflection properties of skin and developed a novel rPPG method based on the model. Further, researchers have utilized face detection and tracking methods such as Bounded Kalman Filter for extracting facial regions of interest \cite{bkf,npjBiases2021}.

More recently, deep learning approaches \cite{eccv2018deepphys,neurips2020multi,cvpr2021dual} have been proposed for the task of physiological estimation. One of the main aspect of DeepPhys \cite{eccv2018deepphys} and MTTS-CAN \cite{neurips2020multi} architectures is the spatial attention mechanism which is used to determine the right regions of interest thereby enabling end-to-end trainability of network. However, all of these methods are evaluated on episodic scores over windowed intervals. To tackle this limitation, the Vision-for-Vitals workshop \cite{revanur2021first}  held at ICCV introduced multiple metrics to promote instantaneous prediction of HR and RR. In our work, we aim to evaluate all methods including ours over these metrics. We also aim to incorporate spatial attention masks and utilize Transformer for temporal learning.

\subsection{Transformers}

A Transformer \cite{vaswani2017attention} consists of a Transformer-encoder and a Transformer-decoder which in turn are composed of several multi-headed self-attention layers. In \cite{vaswani2017attention}, the authors demonstrated high accuracy of Transformer based architecture for multiple language translation tasks. With minor modifications to the proposed Transformer architecture, it has been successfully adapted to a wide range of research problems in computer vision and natural language processing. Particularly, the computer vision community has explored the Transformer based architecture in two forms. 

In one form, the Transformer based architecture includes a convolutional neural network (CNN) backbone that is used as a feature extractor \cite{Lin_2021_CVPR,carion2020end,neimark2021video}. In \cite{carion2020end} the authors employed ImageNet pretrained ResNet based backbone as a spatial feature extractor for the task of object detection using Transformers in an end-to-end manner. In other related work, video Transformers \cite{neimark2021video} have been employed for the goal of temporal modeling of the videos. Here, the convolutional backbone extracts features and the Transformer is used for temporal modelling. In our work, we use a DeepPhys \cite{eccv2018deepphys} based convolutional backbone network and a Transformer for temporal modeling. 

In another form, the architecture is developed purely using Transformer layers. In \cite{dosovitskiy2020image} the authors trained a pure transformer for the task of image classification by dividing the image into multiple parts. In a related work, \cite{yu2021transrppg} aimed at detecting fake 3D printed face masks using Transformer based architecture by drawing motivation from the principles of Photoplethysmography. This is achieved by feeding the Multi-scale Spatio-Temporal maps (MSTmaps) of the facial regions  and background regions along with a positional embedding into a Transformer network.

In this work, we focus on developing a method for the task of instantaneous evaluation of physiological signals by using video Transformer. We aim to train the network in an end-to-end manner by relying on spatial attention module in DeepPhys.

%
%

\section{Methods}

The goal of remote PPG extraction is to effectively extract a bio-signal that contains HR (or RR) using a facial video. To this end, we propose a Transformer-based architecture, inspired by the principles of remote PPG. 
In this section, we first introduce the optical basis of our method by relying on the skin reflection model \cite{wang2016algorithmic,eccv2018deepphys}. Next, we propose the architecture for the HR/RR estimation and finally explain our loss formulation that we used for training the model.

\subsection{Optical basis of video-based bio-signal extraction}

The changes in the volume of blood flowing underlying the facial skin result in subtle color changes. In order to extract this bio-signal, we use the popular skin reflection model that is based on Shafer's dichromatic reflection \cite{wang2016algorithmic,eccv2018deepphys}. At a given time instance $t$ in the video, the reflection of the light back to the camera can be considered as a function that varies in the RGB color space according to the Eq. \ref{eq:ckt}.

\begin{equation} 
        C_k(t)=i(t)\cdot (\mathbf{v}_s(t)+\mathbf{v}_d(t))+\mathbf{v}_n(t)
        \label{eq:ckt}
\end{equation}

Here, $C_k(t)$ is the color intensity of the RGB pixel $k$, $i(t)$ luminance intensity level which is regulated by specular relectance $\mathbf{v}_s$ and diffuse reflectance $\mathbf{v}_d$. The term $\mathbf{v}_n$ is the camera quantization noise. The specular reflection is a mirror-like reflection that bounces the light right off the facial skin while the diffuse component contains useful pulsatile signals. The components $\mathbf{v}_d(t)$ and $\mathbf{v}_s(t)$ can be expressed further in terms of stationary reflection strength, underlying physiological bio-signal $p(t)$ and motion induced changes $m(t)$ (e.g. facial movements, expressions). 

\begin{equation} 
        \mathbf{v}_d(t) = \mathbf{u}_d \cdot d_0 + \mathbf{u}_p \cdot p(t)  
        \label{eqdiff}
\end{equation}

\begin{equation} 
        \mathbf{v}_s(t) = \mathbf{u}_s \cdot (s_0+\Phi(m(t),p(t))) 
        \label{eqspec}
\end{equation}

Here $\mathbf{u}_d$ is the stationary skin reflection strength and $\mathbf{u}_p$ is pulsatile signal strength that varies according to the volume of hemoglobin. Notice how Eq. \ref{eqdiff} does not depend on $m(t)$, while Eq. \ref{eqspec} depends on both $m(t)$ and $p(t)$. Further, $\mathbf{u}_s$ is the unit norm vector indicating the color of light spectrum and $s_0$ is the stationary component of the specular reflection and $\Phi$ is a function of motion $m(t)$ and the physiological signals $p(t)$.

 Next, $i(t)$ can be further decomposed into stationary and varying components according to,

\begin{equation} \label{eqit}
        i(t) = i_0 \cdot (1+\Psi(m(t),p(t))) 
\end{equation}

 Substituting Eq. \ref{eqdiff}, Eq. \ref{eqspec} and Eq. \ref{eqit} into Eq. \ref{eq:ckt}, we obtain Eq. \ref{eqfull}.

\begin{multline} \label{eqfull}
        C_k(t)\approx \mathbf{u}_c \cdot i_0 \cdot c_0+\mathbf{u}_c \cdot i_0 \cdot c_0 \cdot \Psi(m(t),p(t)) + \\
        \mathbf{u}_s \cdot i_0 \cdot \Phi(m(t),p(t))+\mathbf{u}_p \cdot i_0 \cdot p(t)+\mathbf{v}_n(t)
\end{multline}

As a next step, we remove the dominant stationary signal by computing first order derivative of $C_k(t)$ in line with \cite{eccv2018deepphys}. Additionally, we reduce the image size to a size of $36\times 36$ to suppress the quantization noise i.e. $\mathbf{v}_n\approx 0$.

\begin{multline} 
	C'_k(t)\approx \mathbf{u}_c \cdot i_0 \cdot c_0 \cdot (\frac{\partial\Psi}{\partial m}m'(t) + \frac{\partial\Psi}{\partial p}p'(t)) + \\ \mathbf{u}_s \cdot i_0 \cdot (\frac{\partial\Phi}{\partial m}m'(t) 
	+ \frac{\partial\Phi}{\partial p}p'(t))+\mathbf{u}_p \cdot i_0 \cdot p'(t)
\end{multline}

As an additional normalization step, we also divide $C'_k(t)$ by $\mathbf{u}_c \cdot i_0 \cdot c_0$ to remove further stationary luminance intensity across all pixels. In practice, we approximate $C'_k(t)$ by computing the pixel-wise difference for consecutive pairs of frames and normalizing it by using mean frame. We remove the constant factor of 2 from the resulting expression as it is just a scaling term. Therefore, the we obtain $M_k(t)$, which we use for training our model (Eq. \ref{eqinp}).

\begin{equation}
    \label{eqinp}
    M_k(t) = {(C_k(t+1)-C_k(t))}\oslash{(C_k(t+1)+C_k(t))}
\end{equation}

\noindent where $\oslash$ is the Hadamard division (element-wise) operation.

\subsection{Video Transformer for physiological estimation}

\begin{figure}[t]
    \centering
    \includegraphics[width=0.85\columnwidth]{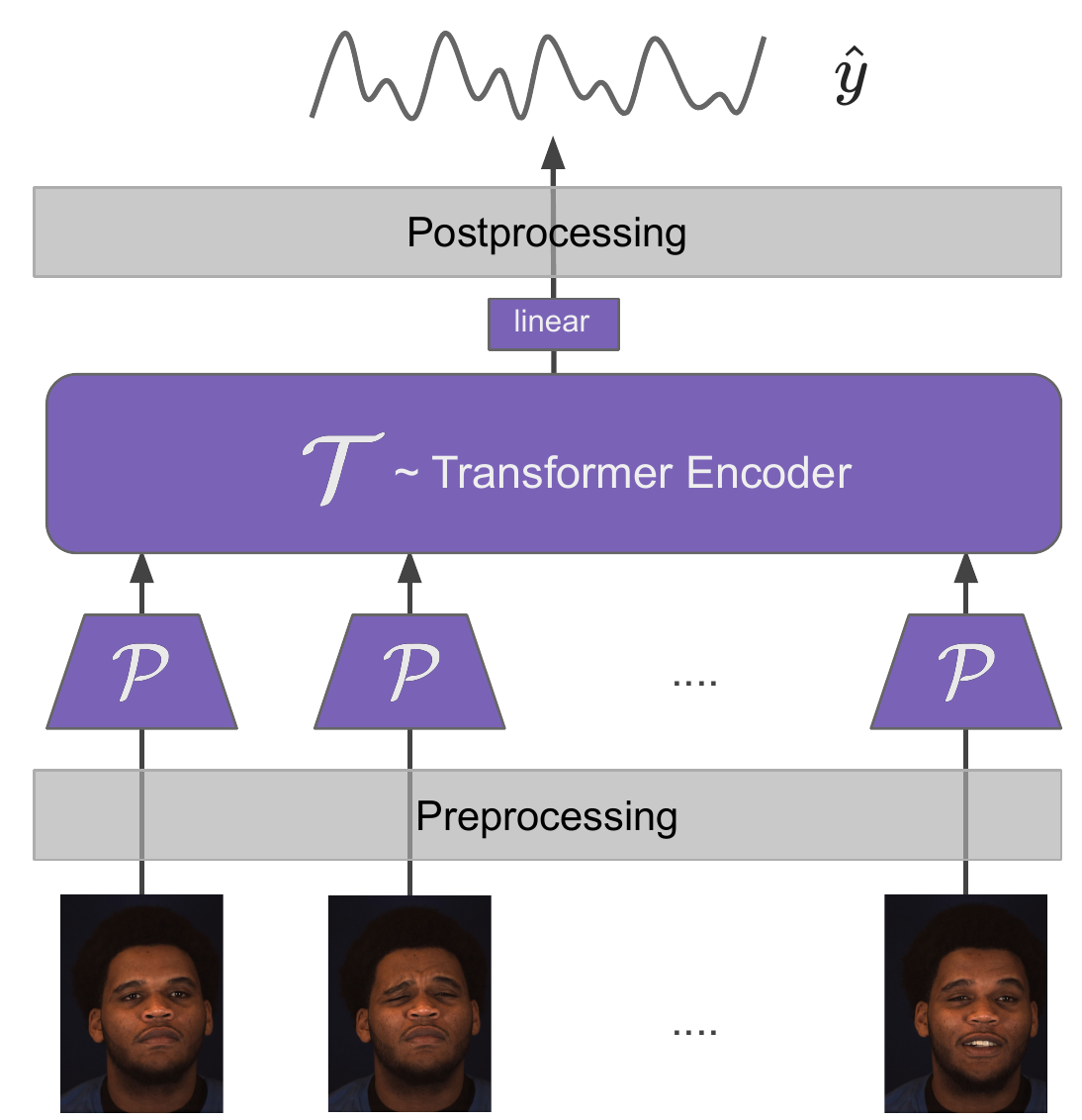}
    \caption{Video transformer for human physiological signal extraction}
    \label{fig:maintrans}
\end{figure}

In this section we propose a Video Transformer for the task of human physiological estimation. As shown in Fig. \ref{fig:maintrans}, the video transformer consists of a spatial backbone $\mathcal{P}$ and frame-level temporal aggregation module $\mathcal{T}$ to assist with learning temporal dependencies of the bio-signal waveform. For the spatial backbone we use the popular DeepPhys-based architecture and for the temporal module, we use the Transformer-encoder architecture. Therefore, our video Transformer is an end-to-end trainable framework unlike the approaches which use facial landmark detection for selecting pixels in the region of interest \cite{cvpr2021dual}.

\begin{figure}[t]
    \centering
    \includegraphics[width=0.85\columnwidth]{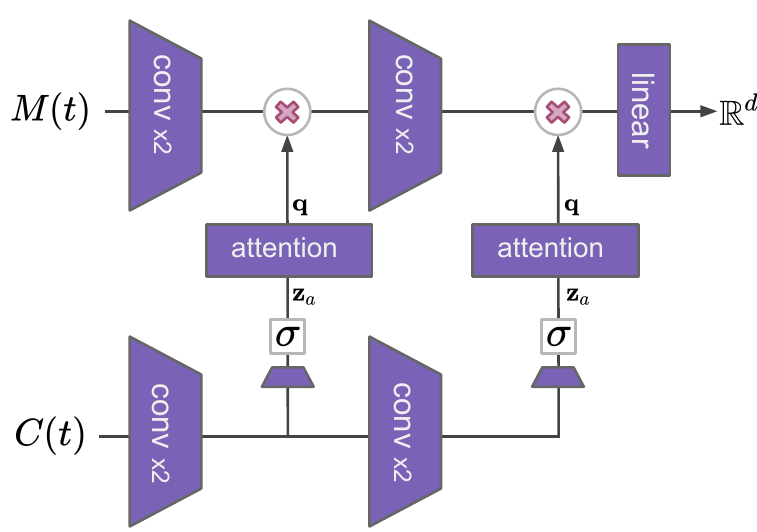}
    \caption{DeepPhys-based encoder $\mathcal{P}$ as a spatial backbone network for video Transformer. Here, $\sigma$ denotes the sigmoid activation and $q$ is the attention mask.}
    \label{fig:maindeepphys}
\end{figure}

\vspace{1mm}
We begin by describing the DeepPhys-based spatial backbone network depicted in Fig. \ref{fig:maindeepphys}. The network consists of two branches for modeling the motion representation and spatial attention. The motivation for using the spatial attention branch is to learn regions of face (via the attention masks) that could assist the motion branch for physiological estimation. The spatial attention branch is trained on a video frame $C_k(t)$ of dimension $36\times36$, and the motion branch is trained on the normalized frame differences $M_k(t)$ of dimension $36\times36$ (Eq. \ref{eqinp}). The attention mask $\mathbf{q}$ is given by,

\begin{equation}
	\mathbf{q} = \frac{(h_{\mathbf{z}} w_{\mathbf{z}})\cdot\mathbf{z}_a}{2\lVert\mathbf{z}_a\rVert_1}
	\label{eqq}
\end{equation}

where $h_{\mathbf{z}}, w_{\mathbf{z}}$ are the dimensions of the input feature map, $\mathbf{z}_a$ are the sigmoid activations of the features from spatial attention branch (Fig. \ref{fig:maindeepphys}). These attention masks are multiplied element-wise across the features of the motion branch. Finally, the FC-layer of $\mathcal{P}$ encodes the spatial features into a vector  of dimension $d$ (Fig. \ref{fig:maindeepphys})


We achieve frame-level temporal aggregation by utilizing a Transformer encoder $\mathcal{T}$ \cite{vaswani2017attention}. The encoder network is trained on the FC-features of $\mathcal{P}$ using $N$ frames of the video where, $N$ is the number of frames used for temporal aggregation. We pass all the features of $\mathcal{P}$ through a linear layer that reduces $d$ dimensional vector into a $d_{\mathcal{T}} = 32$-dimensional embeddings. Along with these embeddings, we additionally include $[CLS]$ token for training the encoder network. Each encoder layer consists of a multi-headed self-attention block with 8 self-attention modules and an MLP layer. Further, at each input step of the Transformer, we also include positional encoding ($PE$) to inject the temporal information into the model. To this end, we utilize the position-encoding layer proposed by \cite{vaswani2017attention} and is controlled according to Eq. \ref{eqpesin} and Eq. \ref{eqpecos}.

\begin{equation}
    \label{eqpesin}
    PE_{(pos,2i)} = \sin(pos / 10000^{2i/d_\mathcal{T}}) 
\end{equation}

\begin{equation}\label{eqpecos}
    PE_{(pos,2i+1)} = \cos(pos / 10000^{2i/d_\mathcal{T}})
\end{equation}

\noindent where, $pos$ is the position and $i$ is the dimension in the position embedding. The output features of the encoder layers are then fed into a single MLP layer to obtain $\hat{y}$.

\subsection{Loss formulation}

We train our model end-to-end using a ground truth signal such as blood pressure (for HR) or respiratory wave (for RR). One straight-forward way to train the model is by using a Mean-Squared Error (MSE) loss. However, MSE loss assumes that the ground truth is accurately synchronized with the bio-signal in the facial video. Unfortunately, it is challenging to perfectly synchronize the bio-signal with ground-truth for two reasons. First the devices used for ground-truth video capture and physiological signal recording are different. Therefore, one has to manually align the video-frames with the ground-truth signal. Second, the ground truth is often collected at a peripheral site such as finger. Therefore, there is additional delay resulting from the Pulse-Transit Time (PTT). A related work \cite{block2020conventional} shows that the time delay for pulse transit between ear and finger is close to 150ms.

One of the other limitations of MSE loss is that it trains the model to learn both amplitude and frequency of a wave. However, for the task of HR/RR estimation, we are interested only in the frequency of the underlying pulsatile signal and not the amplitude of the signal. Therefore, we make use of Maximum Cross-Correlation loss $l(y, \hat{y})$ \cite{gideon2021way} and perform the cross-correlation computation in the frequency domain instead of time domain. 

\begin{equation}
    l(y, \hat{y}) = - c \cdot Max\left( \frac{F^{-1} \{\Omega(F(y) \cdot \overline{F(\hat{y})})\}}{\sigma_y \sigma_{\hat{y}}}  \right)
\end{equation}

where $\hat{y}$ is the ground-truth as computed by signal differences $\Delta p$ and $y$ is the predicted waveform. Further, $\Omega$ is a bandpass operator which retains only frequencies of interest, $c$ is the ratio of power present inside the frequency range of heart rate to the total power, $F$ is the Fourier-transform operator and $(\overline{\cdot})$ is the conjugate operator.

\section{Results}

\subsection{Implementation details}
We reduced the input image size to $36\times36$ inline with \cite{eccv2018deepphys} and computed the normalized input frame difference for motion branch. For training the video Transformer, we fixed the number of input frames to $N$ (where $N=100$ for HR estimation and $N=1000$ for RR estimation) and trained our network end-to-end. During inference, we used the predictions from $\mathcal{T}$ and computed cumulative sum to obtain the final waveform prediction. After that, we calculated the Fourier Transform of the waveform and applied bandpass filter to limit the frequencies within the range of HR / RR and obtained $\hat{y}$. For HR model, we used $d=128$ and for RR model, we used $d=32$.

\begin{figure}[t]
    
    \centering
    \includegraphics[width=0.8\columnwidth]{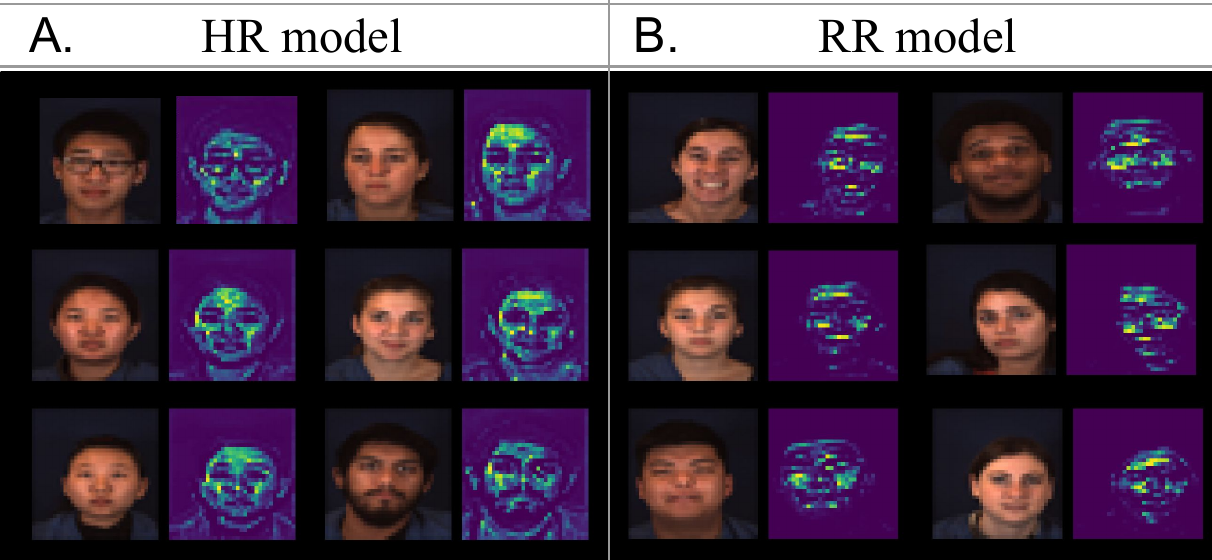}
    \caption{Spatial attention masks obtained for the HR model (left) and the RR model (right) on V4V test set. }
    \label{fig:hr_rr}
\end{figure}

\subsection{Datasets and evaluation protocol}

We use Vision-for-Vitals (V4V) dataset that consists of 179 subjects and 1358 videos in total. 
The V4V dataset contains continuous blood pressure waveform recorded at 1KHz, frame-aligned HR and frame-aligned RR. We use the V4V training dataset for training the model and report the performance of our model on both V4V validation set and V4V test set. We follow the evaluation protocol set forth in the V4V challenge \cite{revanur2021first} and report continuous MAE ($cMAE$) and continuous RMSE ($cRMSE$). 

\begin{equation}
    cMAE = \frac{\Sigma_i |\widehat{HR}_i - HR_i|}{N'}
\end{equation}

\begin{equation}
    cRMSE=\sqrt{\frac{(\Sigma_i|\widehat{HR}_i -HR_i|^2)}{N'}}
\end{equation}

\noindent where, $\widehat{HR}_i$ and $HR_i$ are the predicted HR and ground-truth HR for the frame $i$ respectively and $N'$ is total number of frames in test-set. We use the same evaluation protocol for benchmarking RR results. Further, in order to enable evaluation of all methods on continuous evaluation protocol, we use a short moving window over the predicted blood volume pulse for HR (and predicted respiratory wave for RR) and employed FFT to predict continuous HR.

\subsection{Heart rate estimation results}

Table \ref{tab:hr} shows the comparison of our method against traditional non-deep learning methods (implemented in \cite{mcduff2019iphys}) - Green \cite{green2008remote}, POS \cite{wang2016algorithmic} and recent deep-learning methods - DeepPhys \cite{eccv2018deepphys} and TS-CAN \cite{neurips2020multi}. It is important to note, that for fair comparison, we excluded studies that either utilize external training data \cite{brian2021hybrid} or access the test set for domain adaptation \cite{shisen2021selfsup} on this benchmark. For computing the HR, we used a bandpass filter with range of $[0.7, 2.5]$.

\begin{table}[t]
    \centering
    
    \caption{Comparison of our method against previous works for HR estimation on the V4V validation set and V4V test set. Note that lower $cMAE$ and lower $cRMSE$ are better ($\downarrow$). }
    \begin{tabular}{|p{80pt}|c|c|c|c|c|}
    \hline
    \multirow{2}{*}{Name} & \multirow{2}{*}{ \makecell{$cMAE$ \\ Val. set ($\downarrow$)}}  & \multirow{2}{*}{ \makecell{$cRMSE$ \\ Val. set ($\downarrow$)}} & \multirow{2}{*}{\makecell{$cMAE$ \\ Test set ($\downarrow$)}} & \multirow{2}{*}{\makecell{$cRMSE$ \\ Test set ($\downarrow$)}} \\ 
    & & & & \\
    \hline
    Green \cite{green2008remote}      & 16.5          & 21.4 & 15.5          & 21.9 \\
    POS \cite{wang2016algorithmic}    & 17.3          & 21.2     & 15.3          & 21.8 \\
    ICA \cite{ica2010non}             & 13.9          & 20.0 & 15.1      & 20.6 \\
    DeepPhys \cite{eccv2018deepphys}  & 13.6          &  18.1    & 14.7          & 19.7 \\
    TS-CAN \cite{neurips2020multi}    & 11.7          & 17.8     & 13.9          & 19.2\\
    Ours                              & \textbf{10.3} &\textbf{ 16.1 }    & \textbf{13.0} & \textbf{18.8} \\

    \hline
    \end{tabular}
    \label{tab:hr}

\end{table}

\begin{table}[t]
    \centering
    
    \caption{Comparison of our method against previous works for RR estimation on the V4V validation set and V4V test set. Note that lower $cMAE$ and lower $cRMSE$ are better ($\downarrow$). }
    \begin{tabular}{|p{80pt}|c|c|c|c|}
    \hline
\multirow{2}{*}{Name} & \multirow{2}{*}{ \makecell{$cMAE$ \\ Val. set ($\downarrow$)}}  & \multirow{2}{*}{ \makecell{$cRMSE$ \\ Val. set ($\downarrow$)}} & \multirow{2}{*}{\makecell{$cMAE$ \\ Test set ($\downarrow$)}} & \multirow{2}{*}{\makecell{$cRMSE$ \\ Test set ($\downarrow$)}} \\ 
    & & & & \\
    \hline
    Green \cite{green2008remote}     & 5.9          & 6.8    & 7.0          &  7.5         \\
    POS \cite{wang2016algorithmic}   & 6.1          & 6.9    & 6.5          &  6.9         \\
    ICA \cite{ica2010non}            & 6.4          & 7.2    & 5.8          &  6.2         \\
    DeepPhys \cite{eccv2018deepphys} & 5.0          & 6.1    & 5.5          & \textbf{5.9}          \\
    Ours                             & \textbf{4.8} & \textbf{5.6}    & \textbf{5.4} &  6.0         \\
    \hline
    \end{tabular}
    \label{tab:rr}
    
\end{table}

\subsection{Spatial attention mask}

The spatial attention mask offers visibility into where the model is extracting the HR and RR in a given frame. As shown in Fig. \ref{fig:hr_rr}, the base encoder is able to focus on the regions corresponding to facial skin for extraction of physiological signals. Notice how the model is excluding facial accessories such as eye-glasses (top-left subject in Fig. \ref{fig:hr_rr}A) and facial hair (bottom-right subject in Fig. \ref{fig:hr_rr}A).

\subsection{Respiration rate estimation results}

We train our proposed model on the continuous respiration waveform of the V4V training dataset and we report the results on the V4V validation set and the V4V test set in Table \ref{tab:rr}. We compare our results with traditional and deep-learning based approaches. For computing the RR, we extracted the biosignal \cite{tarassenko2014non3in1} and used a bandpass filter with range of $[0.13, 0.34]$. Results indicate that our method performs better than the other approaches on the V4V dataset.

\section{Conclusion}

In this paper, we take a step towards instantaneous prediction of physiological signals by utilizing a Transformer based architecture for extracting the heart rate and respiration rate from facial videos. We train the video Transformer model in an end-to-end manner using a cross-correlation loss in the frequency domain. The results of our approach over continuous evaluation metric using Vision-for-Vitals (V4V) dataset shows that the model is able to outperform both shallow and deep learning methods on the task of heart rate and respiration rate estimation. As part of future work Video Transformers can be used to tackle the domain shift problem (laboratory to real-world) \cite{icassp_da_2020,kundu2020towards} and can be used to extract other physiological signals such as Oxygen Saturation (SpO$_2$) \cite{tarassenko2014non3in1,acs2021}.

\section*{Acknowledgement}

\noindent This project is funded by the Bill \& Melinda Gates Foundation (BMGF). Any opinions, findings, or conclusions are those of the authors and do not necessarily reflect the views of the sponsors.

%
%
%

%
%
%
\bibliographystyle{splncs04}
\bibliography{main}

\begin{thebibliography}{10}
\providecommand{\url}[1]{\texttt{#1}}
\providecommand{\urlprefix}{URL }
\providecommand{\doi}[1]{https://doi.org/#1}

\bibitem{block2020conventional}
Block, R.C., Yavarimanesh, M., Natarajan, K., Carek, A., Mousavi, A.,
  Chandrasekhar, A., Kim, C.S., Zhu, J., Schifitto, G., Mestha, L.K., et~al.:
  Conventional pulse transit times as markers of blood pressure changes in
  humans. Scientific Reports  \textbf{10}(1) (2020)

\bibitem{carion2020end}
Carion, N., Massa, F., Synnaeve, G., Usunier, N., Kirillov, A., Zagoruyko, S.:
  End-to-end object detection with transformers. In: European Conference on
  Computer Vision (2020)

\bibitem{eccv2018deepphys}
Chen, W., McDuff, D.: Deepphys: Video-based physiological measurement using
  convolutional attention networks. In: Proceedings of the European Conference
  on Computer Vision (ECCV) (2018)

\bibitem{npjBiases2021}
Dasari, A., Prakash, S.K.A., Jeni, L.A., Tucker, C.: Evaluation of biases in
  remote photoplethysmography methods. NPJ Digital Medicene  (2021)

\bibitem{chrom2013robust}
De~Haan, G., Jeanne, V.: Robust pulse rate from chrominance-based rppg. IEEE
  Transactions on Biomedical Engineering  \textbf{60}(10) (2013)

\bibitem{dosovitskiy2020image}
Dosovitskiy, A., Beyer, L., Kolesnikov, A., Weissenborn, D., Zhai, X.,
  Unterthiner, T., Dehghani, M., Minderer, M., Heigold, G., Gelly, S., et~al.:
  An image is worth 16x16 words: Transformers for image recognition at scale.
  arXiv preprint arXiv:2010.11929  (2020)

\bibitem{gideon2021way}
Gideon, J., Stent, S.: The way to my heart is through contrastive learning:
  Remote photoplethysmography from unlabelled video. In: Proceedings of the
  IEEE/CVF International Conference on Computer Vision (2021)

\bibitem{brian2021hybrid}
Hill, B., Liu, X., McDuff, D.: Beat-to-beat cardiac pulse rate measurement from
  video. In: Proceedings of the IEEE/CVF International Conference on Computer
  Vision Workshops (2021)

\bibitem{kundu2020towards}
Kundu, J.N., Venkat, N., Revanur, A., Babu, R.V., et~al.: Towards inheritable
  models for open-set domain adaptation. In: Conference on Computer Vision and
  Pattern Recognition (CVPR) (2020)

\bibitem{icassp_da_2020}
Lee, E., Ho, A., Wang, Y.T., Huang, C.H., Lee, C.Y.: Cross-domain adaptation
  for biometric identification using photoplethysmogram. In: ICASSP 2020 - 2020
  IEEE International Conference on Acoustics, Speech and Signal Processing
  (ICASSP) (2020). \doi{10.1109/ICASSP40776.2020.9053604}

\bibitem{Lin_2021_CVPR}
Lin, K., Wang, L., Liu, Z.: End-to-end human pose and mesh reconstruction with
  transformers. In: Proceedings of the IEEE/CVF Conference on Computer Vision
  and Pattern Recognition (CVPR). pp. 1954--1963 (June 2021)

\bibitem{neurips2020multi}
Liu, X., Fromm, J., Patel, S., McDuff, D.: Multi-task temporal shift attention
  networks for on-device contactless vitals measurement. arXiv preprint
  arXiv:2006.03790  (2020)

\bibitem{cvpr2021dual}
Lu, H., Han, H., Zhou, S.K.: Dual-gan: Joint bvp and noise modeling for remote
  physiological measurement. In: Proceedings of the IEEE/CVF Conference on
  Computer Vision and Pattern Recognition (2021)

\bibitem{mcduff2019iphys}
McDuff, D., Blackford, E.: iphys: An open non-contact imaging-based
  physiological measurement toolbox. In: 2019 41st Annual International
  Conference of the IEEE Engineering in Medicine and Biology Society (EMBC).
  IEEE (2019)

\bibitem{neimark2021video}
Neimark, D., Bar, O., Zohar, M., Asselmann, D.: Video transformer network.
  arXiv preprint arXiv:2102.00719  (2021)

\bibitem{eccv_20_video}
Niu, X., Yu, Z., Han, H., Li, X., Shan, S., Zhao, G.: Video-based remote
  physiological measurement via cross-verified feature disentangling. In:
  European Conference on Computer Vision (2020)

\bibitem{npjafib2020}
Pereira, T., Tran, N., Gadhoumi, K., M.~Pelter, M., Do, D.H., Lee, R.J.,
  Colorado, R., Meisel, K., Hu, X.: Photoplethysmography based atrial
  fibrillation detection: a review. NPJ Digital Medicene  (2020)

\bibitem{ica2010non}
Poh, M.Z., McDuff, D.J., Picard, R.W.: Non-contact, automated cardiac pulse
  measurements using video imaging and blind source separation. Optics express
  \textbf{18}(10) (2010)

\bibitem{bkf}
Prakash, S.K.A., Tucker, C.S.: Bounded kalman filter method for motion-robust,
  non-contact heart rate estimation. Biomedical optics express  \textbf{9}(2)
  (2018)

\bibitem{revanur2021first}
Revanur, A., Li, Z., Ciftci, U.A., Yin, L., Jeni, L.A.: The first vision for
  vitals (v4v) challenge for non-contact video-based physiological estimation.
  In: Proceedings of the IEEE/CVF International Conference on Computer Vision
  Workshops (2021)

\bibitem{acs2021}
Shao, D., Liu, C., Tsow, F.: Noncontact physiological measurement using a
  camera: A technical review and future directions. ACS Sensors  \textbf{6}(2),
   321--334 (2021). \doi{10.1021/acssensors.0c02042},
  \url{https://doi.org/10.1021/acssensors.0c02042}, pMID: 33434004

\bibitem{shisen2021selfsup}
Stent, S., Gideon, J.: Estimating heart rate from unlabelled video. In:
  Proceedings of the IEEE/CVF International Conference on Computer Vision
  Workshops (2021)

\bibitem{tarassenko2014non3in1}
Tarassenko, L., Villarroel, M., Guazzi, A., Jorge, J., Clifton, D., Pugh, C.:
  Non-contact video-based vital sign monitoring using ambient light and
  auto-regressive models. Physiological measurement  \textbf{35}(5) (2014)

\bibitem{vaswani2017attention}
Vaswani, A., Shazeer, N., Parmar, N., Uszkoreit, J., Jones, L., Gomez, A.N.,
  Kaiser, {\L}., Polosukhin, I.: Attention is all you need. In: Advances in
  neural information processing systems. pp. 5998--6008 (2017)

\bibitem{green2008remote}
Verkruysse, W., Svaasand, L.O., Nelson, J.S.: Remote plethysmographic imaging
  using ambient light. Optics express  \textbf{16}(26) (2008)

\bibitem{wang2016algorithmic}
Wang, W., den Brinker, A.C., Stuijk, S., De~Haan, G.: Algorithmic principles of
  remote ppg. IEEE Transactions on Biomedical Engineering  \textbf{64}(7)
  (2016)

\bibitem{Wu12Eulerian}
Wu, H.Y., Rubinstein, M., Shih, E., Guttag, J., Durand, F., Freeman, W.T.:
  Eulerian video magnification for revealing subtle changes in the world. ACM
  Transactions on Graphics (Proc. SIGGRAPH 2012)  \textbf{31}(4) (2012)

\bibitem{yu2021transrppg}
Yu, Z., Li, X., Wang, P., Zhao, G.: Transrppg: Remote photoplethysmography
  transformer for 3d mask face presentation attack detection. IEEE Signal
  Processing Letters  (2021)

\end{thebibliography}

\end{document}